\documentclass[letterpaper]{article} 
\usepackage{aaai25}  
\usepackage{times}  
\usepackage{helvet}  
\usepackage{courier}  
\usepackage[hyphens]{url}  
\usepackage{graphicx} 
\usepackage{natbib}  
\usepackage{amsthm}
\usepackage{caption} 
\usepackage{algorithmic}
\usepackage[ruled,linesnumbered]{algorithm2e}

\usepackage{subcaption}

\graphicspath{ {./images/} }

\usepackage{amsmath,amsfonts,bm}

\def\eqref#1{equation~\ref{#1}}

\def\1{\bm{1}}

\def\rvepsilon{{\mathbf{\epsilon}}}

\def\vp{{\bm{p}}}

\def\vx{{\bm{x}}}
\def\vy{{\bm{y}}}
\def\vz{{\bm{z}}}

\def\mI{{\bm{I}}}

\def\mW{{\bm{W}}}

\DeclareMathAlphabet{\mathsfit}{\encodingdefault}{\sfdefault}{m}{sl}
\SetMathAlphabet{\mathsfit}{bold}{\encodingdefault}{\sfdefault}{bx}{n}
\newcommand{\tens}[1]{\bm{\mathsfit{#1}}}

\def\tZ{{\tens{Z}}}

\def\gD{{\mathcal{D}}}

\def\gN{{\mathcal{N}}}

\def\gT{{\mathcal{T}}}

\newcommand{\E}{\mathbb{E}}
\newcommand{\Ls}{\mathcal{L}}

\DeclareMathOperator*{\argmax}{arg\,max}

\title{Single Domain Generalization with Adversarial Memory}
\author{
 Hao Yan,\; Marzi Heidari,\;  Yuhong Guo
}
\affiliations{
 School of Computer Science, Carleton University, Ottawa, Canada\\

 \{haoyan6@cmail., marziheidari@cmail., yuhong.guo@\}carleton.ca
}

\usepackage{bibentry}

\begin{document}

\maketitle

\begin{abstract}

Domain Generalization (DG) aims to train models that can generalize to unseen testing domains by leveraging data from multiple training domains. However, traditional DG methods rely on the availability of multiple diverse training domains, limiting their applicability in data-constrained scenarios. Single Domain Generalization (SDG) addresses the more realistic and challenging setting by restricting the training data to a single domain distribution.
The main challenges in SDG stem from the limited diversity of training data and the inaccessibility of unseen testing data distributions. To tackle these challenges, we propose a single domain generalization method that leverages an adversarial memory bank to augment training features. 
Our memory-based feature augmentation network maps both training and testing features into an invariant subspace spanned by diverse memory features, implicitly aligning the training and testing domains in the projected space.
To maintain a diverse and representative feature memory bank, we introduce an adversarial feature generation method that creates features extending beyond the training domain distribution. 
Experimental results demonstrate that our approach achieves state-of-the-art performance on standard single domain generalization benchmarks.

\end{abstract}

\section{Introduction}
Traditional deep classification models are typically trained under the assumption that the training and testing data share the same distribution. However, this assumption limits their effectiveness in cross-domain applications. Domain Generalization (DG) has emerged as a solution to this issue, aiming to train deep classification models capable of generalizing to unseen testing domains using data from multiple training domains.
Typical DG methods focus on either domain-invariant representation learning \citep{li2018deep}, or meta-learning \citep{du2020learning} to simulate domain shifts within the training domains. Other approaches utilize ensemble learning \citep{arpit2021ensemble} to enhance testing performance. More recent works \citep{foret2020sharpness, cha2021swad, zhang2023flatness} emphasize the importance of optimizing the flatness of empirical risk in relation to model parameters to improve generalization. However, the traditional DG problem setting relies on access to multiple training domains, limiting its applicability in scenarios where the availability of diverse training data is constrained.

To address this limitation, Single Domain Generalization (SDG) focuses on training models using data from a single training domain and evaluates their performance across multiple unseen testing domains. This setting presents a more challenging yet realistic problem, as it must contend with the limited diversity of the training data and the inaccessibility of unseen testing data distributions. The former challenge restricts the direct application of traditional DG methods, which rely on multiple training domains to analyze domain shifts, while the latter challenge raises the bar for the model's generalization ability.
Recent works addressing the SDG setting predominantly employ various data augmentation techniques to expand the data distribution within the single training domain. For instance, \citet{Long2020Maximum} augmented images directly in the pixel space to simulate worst-case domain shifts, \citet{Cugu_2022_CVPR} applied visual corruption transformations to mimic different training domains, and \citet{xu2023simde} used simulated domain variability to enhance the model's robustness. However, these methods empirically expand the training data distribution in random directions rather than uniformly broadening the boundaries, leading to inconsistent performance improvements across different unseen testing domains. Some domains see significant gains, while others see minimal benefits.

In this paper, we propose a novel method, Single Domain Generalization with Adversarial Memory (SDGAM), that leverages adversarial memory for single domain generalization. Our approach employs a maintained feature memory bank to augment training features. Specifically, the proposed memory-based feature augmentation network maps training features into an invariant subspace spanned by diverse feature vectors within the memory bank. This invariant subspace mapping implicitly aligns the training and testing feature distributions in the projected space without requiring access to testing data during the training phase.
To maintain a diverse and representative feature memory bank, we carefully design the feature memory initialization and updating strategies. Our proposed adversarial feature generation method uses noisy gradients to update training features, generating features that extend beyond the training domain distribution. These generated adversarial features are then used to update the memory bank, ensuring it remains representative and up-to-date.
The classification model and feature augmentation network trained on the training data are utilized for inference on testing data. As the maintained memory bank is relatively small, the additional computational cost introduced by the feature augmentation network is negligible.
To evaluate the effectiveness of our SDGAM method, we conduct experiments under the standard single domain generalization setting. Our method achieves state-of-the-art performance on standard single domain generalization benchmarks.

\section{Related Work}

\subsection{Domain Generalization}
Domain Generalization (DG) aims to train models using data from multiple training domains and evaluate its performance on data from unseen testing domain. 
Domain-invariant representation learning-based DG methods \citep{li2018deep, akuzawa2020adversarial, mahajan2021domain} focus on learning domain-invariant models through adversarial domain alignment or by minimizing criteria that measure domain discrepancies.
Meta-learning-based DG methods \citep{li2018learning, du2020learning} simulate domain shifts during training by constructing meta-train and meta-test sets, aiming to improve performance on both during each optimization step. 
Ensemble-learning-based DG methods \citep{zhou2021domain, arpit2021ensemble} train individual models on data from each training domain and then use ensemble networks to combine multiple model predictions.
Flatness-aware DG methods \citep{foret2020sharpness, cha2021swad, zhang2023flatness} seek flat minima of empirical risk in relation to model parameters to enhance generalization capability. 
However, most of these DG methods require access to data from multiple training domains, which limits their direct applicability in a single domain generalization setting.

\subsection{Single Domain Generalization}
Single Domain Generalization (SDG) targets the intricate challenge of generalizing models trained exclusively on a single training domain without exposure to testing domain distributions. This approach differs from traditional domain generalization, which utilizes multiple training domains to bolster model robustness. SDG focuses on enhancing the training data to anticipate and adapt to potential shifts encountered in novel, unseen testing domains. 

Current SDG methodologies are categorized into three main streams. The first category employs traditional data augmentation techniques to fortify in-domain robustness, though their efficacy in bridging substantial domain discrepancies is often limited. Noteworthy contributions in this area include enhanced augmentation strategies \citep{devries2017improved}, AugMix \citep{hendrycks2019augmix}, and AutoAugment \citep{Cubuk_2020_CVPR_Workshops}, which primarily augment sample variability within the training domain but fall short in out-of-domain generalization.  \citet{lian2021geometry} expanded this approach by integrating geometric transformations to increase training sample diversity, albeit with moderate success in addressing domain shifts. ACVC \citep{Cugu_2022_CVPR} adopts visual corruption transformations to emulate distinct training domains and maintains visual attention consistency between original and altered samples.
The second category introduces adversarial data augmentation techniques, which ingeniously manipulate either pixel space or latent features to emulate domain variability. Pioneering works by \citet{volpi2018generalizing} and \citet{Long2020Maximum} involve augmenting images directly in the pixel space to generate worst-case domain shifts. \citet{zhang2023adversarial} advanced this notion by perturbing latent feature statistics, although producing extensive domain shifts consistently remains a challenge. Adversarial AutoAugment \citep{zhang2019adversarial} devises augmentation policies through adversarial learning, significantly improving in-domain generalization.
The third and final category capitalizes on generative models to synthesize training data. Innovations by \citet{qiao2020learning},  \citet{wang2021learning}, and  \citet{li2021progressive} utilize generative adversarial networks (GANs) and variational autoencoders (VAEs) to create diverse, though domain-constrained, samples. MCL \citep{chen2023meta} first emulates domain shifts via an auxiliary testing domain, dissects their origins, and subsequently mitigates these shifts for model adaptation. SimDE \citep{xu2023simde} broadens the training domain to emulate multiple domains during training, thereby enhancing the model’s robustness and generalizability across unseen domains by exploiting simulated domain variability.

The previous study in SDG is primarily restricted by their reliance on random in-domain data augmentation, which may not adequately simulate the variability and shifts found in unseen domains. Our proposed SDGAM differs significantly by integrating a memory-based feature augmentation mechanism that spans the features into an invariant subspace beyond the training domain's distribution.

\subsection{Memory-Based Approaches}
Memory-based approaches have markedly advanced the field of image generalization, leveraging historical data to bolster models’ predictive accuracy and adaptability across diverse tasks and environments. Foundational advancements by Memory Networks \citep{weston2015memory} integrated memory modules with learning algorithms, significantly enhancing prediction capabilities. This concept has been further refined in constructs like the Differentiable Neural Computer, which merges neural networks with dynamic external memory, facilitating sophisticated data integration \citep{santoro2016meta}.

In the context of domain generalization, memory-based strategies employ mechanisms that encode domain-specific knowledge to heighten adaptability when faced with novel domains \citep{zhao2021learning}. Specifically, STEAM \citep{chen2021style} introduces a dual-memory architecture, distinctively encoding style and semantic attributes, thereby robustly addressing challenges in domain generalization.

\section{Proposed Method}

In this section, we introduce our proposed method, Single Domain Generalization with Adversarial Memory (SDGAM). We propose a memory-based feature augmentation network that maps features to an invariant subspace using a diverse memory bank. To construct this diverse memory bank, we employ an adversarial feature generation method designed to create features that extend beyond the training domain distribution. The specifics of this approach are further elaborated in the following sections.

\paragraph{Problem Setting}
We address the challenging problem of Single Domain Generalization (SDG), where only data from a single training domain are available, and the model is evaluated on multiple unseen testing domains. Let the data distribution of the single training domain be denoted as $\gD$, 
with a dataset $D$ sampled from $\gD$ as the training set.
We focus on a typical image classification task, where the classification model consists of a feature encoder $f_\theta$ parametrized by $\theta$ and a linear classification head $h_\phi$ parametrized by $\phi$. Each element in $D$ is denoted as $(\vx, \vy) \in D$, where $\vx$ represents the image data and $\vy$ is the one-hot label vector of length $N_c$ that denotes the number of classes.
Given an instance $\vx$, the feature encoder $f_\theta$ extracts a feature vector $\vz$ from $\vx$, i.e., $\vz = f_\theta(\vx)$.
The classification model is trained on the training set $D$ and evaluated on multiple unseen testing sets $T = \{T_i\}_{i=1}^{N_T}$, where $N_T$ denotes the number of testing sets. Each testing set $T_i$ is uniformly sampled from its corresponding testing domain distribution $\gT_i$. 

\subsection{Memory-based Feature Augmentation}

The traditional Empirical Risk Minimization (ERM) method trains the classification model on the training set $D$ using cross-entropy loss. However, models trained with ERM often struggle to generalize to unseen testing domains, primarily due to cross-domain distribution discrepancies and the inaccessibility of testing domain data. To address this, we propose a feature augmentation network that maps the extracted features to an invariant subspace using a diverse feature memory bank. This feature augmentation network implicitly aligns the training and testing feature distributions without requiring access to the testing data.

Specifically, given a training instance $\vx \in D$, the feature encoder $f_\theta$ extracts a feature vector $\vz = f_\theta(\vx)$ with a length of $d_\vz$. We maintain an updated feature memory bank $\tZ$, 
which is obtained from historical training features:
\begin{equation}
    \tZ = \{\vz_1, \cdots, \vz_{N_m}\}
\end{equation}
where $N_m$ denotes the size of the memory bank, and $\vz_i$ represents each individual feature vector in the memory bank. The initialization and updating process of the feature memory bank will be discussed in the next section.

We use attention mechanism \citep{Attention2017} to measure the similarity between the extracted feature $\vz$ and each feature vector $\vz_i$ in the memory bank $\tZ$. Specifically,
\begin{equation}
	\alpha_i = \frac{\exp((\mW_q \vz)^\top (\mW_k \vz_i) / \sqrt{d_h})}{\sum_{j=1}^{N_m} \exp((\mW_q \vz)^\top (\mW_k \vz_j) / \sqrt{d_h})}
\end{equation}
where $\mW_q$ and $\mW_k$ are the learnable query and key projection matrices, both with dimensions $d_h \times d_\vz$, and $d_h$ denotes the hidden dimension of the projected features. The calculated $\alpha_i$ measures the similarity between $\vz$ and $\vz_i$ in the projected hidden space, and the scaling factor $\frac{1}{\sqrt{d_h}}$ prevents the loss function from producing extremely small gradients with respect to the projection matrices \citep{Attention2017}.

We use the calculated similarities $\{\alpha_i\}_{i=1}^{N_m}$ as weights to linearly combine the feature vectors $\{\vz_i\}_{i=1}^{N_m}$ in the feature memory bank. Specifically,
\begin{equation}
    \vz^a = \sum_{i=1}^{N_m} \alpha_i \vz_i
\end{equation}
Through this attention mechanism, the original feature vector $\vz$ is mapped into a new feature vector $\vz^a$ using the feature vectors in the memory bank and their similarities with $\vz$. 
These memory bank vectors in $\tZ$ serve as basis vectors that describe the original vector $\vz$ in the linearly spanned subspace, $\text{span}(\tZ)$.
We refer to the resulting feature vector $\vz^a$ as the augmenting feature and denote the feature augmentation network as $g$, i.e.,
\begin{equation}
    \vz^a = g(\vz; \mW_q, \mW_k, \tZ)
\end{equation}
Considering the domain discrepancy between the training and testing domains, our proposed feature augmentation network $g$ can map both training and testing features into the same subspace, $\text{span}(\tZ)$, thereby implicitly reducing the domain discrepancy.

Instead of replacing the original feature vector $\vz$ with the augmenting feature vector $\vz^a$ for classification, we augment the original feature vector $\vz$ by concatenating $\vz^a$ after $\vz$ along the feature dimension to create a longer feature vector $\vz' = [\vz; \vz^a]$ with a length of $2 \times d_\vz$. Consequently, the number of input neurons in the linear classification head $h$ is doubled to $2 \times d_\vz$.
We refer to this longer vector $\vz'$ as the augmented feature vector, which is then used for classification model training with the following cross-entropy loss:
\begin{equation}\label{eq:loss-aug}
    \Ls_\text{aug} = \E_{(\vx,y)\in D} [\ell_{ce}(h_\phi([f_\theta(\vx); g({f}_{\bar{\theta}}(\vx))]), \vy')]
\end{equation}
Here, $\bar{\theta}$ indicates that the gradient of the loss function is stopped from back-propagating through the feature augmentation network $g$ to the feature encoder $f_\theta$. This is done to prevent unstable gradients from affecting the parameters of $f_\theta$ due to periodic memory updates.
The label vector for the augmented feature vector $\vz'$ is denoted as $\vy'$, and it is calculated as follows:
\begin{equation}\label{eq:label-mixup}
    \vy' = \beta \vy + (1-\beta) \sum_{i=1}^{N_m} \alpha_i \vy_i
\end{equation}

The reasoning behind the calculation of $\vy'$ is as follows. Let the weight matrix of the linear classification head $h$ be denoted as $\mW$, with dimensions $N_c \times 2d_\vz$, where $N_c$ is the number of classes. Ignoring the bias vector, the predicted logits can be expressed as:
\begin{equation}\label{eq:logits-ensemble}
    h_\phi([\vz; \vz^a]) = \mW^{N_c\times 2d_\vz}
    \begin{bmatrix}
        \vz \\
        \vz^a
    \end{bmatrix}
    = \mW_1^{N_c \times d_\vz} \vz + \mW_2^{N_c \times d_\vz} \vz^a
\end{equation}
where $\mW^{N_c \times 2d_\vz} = [\mW_1^{N_c \times d_\vz}, \mW_2^{N_c \times d_\vz}]$. It can be observed that the predicted logits of the concatenated feature vector are the sum of two logits vectors: one predicted by the model $\mW_1$ using the feature $\vz$, and the other predicted by the model $\mW_2$ using the feature $\vz^a$. 
On the other hand, the feature vector $\vz$ is associated with the one-hot label vector $\vy$. The label vector for the feature $\vz^a$ can be expressed as $\sum_{i=1}^{N_m} \alpha_i \vy_i$, 
where $\vy_i$ is the label vector for the memory feature vector $\vz_i$.
This can be seen as a feature-level MixUp \citep{zhang2018mixup} of the memory feature vectors and their corresponding label vectors. 
Linearly combining the two label vectors of $\vz$ and $\vz^a$ provides a reasonable label vector $\vy'$ for the augmented feature vector $\vz'$, with $\beta$ controlling the importance of the original feature vector $\vz$ to the logits ensemble.
Moreover, the implicit logits-level ensemble can be expected to further improve the model's performance, validating the choice of feature concatenation over simply replacing the original feature vector with the augmenting feature vector for classification.

In addition to the training loss function in Eq~(\ref{eq:loss-aug}), which uses the augmented feature vector, we introduce another cross-entropy loss that uses the original feature vector without augmentation. 
Since the number of input neurons in the classification head $h$ has been expanded from $d_\vz$ to $2d_\vz$, we simply duplicate $\vz$ for feature concatenation to match the input size of the classification head. Specifically,
\begin{equation}\label{eq:loss-cls}
    \Ls_\text{cls} = \E_{(\vx,y)\in D} [\ell_{ce}(h_\phi([f_\theta(\vx); f_\theta(\vx)]), \vy)]
\end{equation}
where $\vy$ is the label vector of the input data $\vx$. 
This feature duplication and concatenation can also be analyzed using logits-level ensemble. By replacing $\vz^a$ in Eq~(\ref{eq:logits-ensemble}) with $\vz$, the predicted logits $h_\phi([\vz; \vz])$ can be seen as the equal average of two logits vectors predicted by two separate models, $\mW_1$ and $\mW_2$, using same feature $\vz$. 
Since $\mW_1$ and $\mW_2$ are randomly initialized with different values, this feature duplication and concatenation can be viewed as a logits-level ensemble of predictions from two separate classification heads, which similarly improves the model's performance.
Therefore, the overall training loss is formulated as follows:
\begin{equation}\label{eq:overall-loss}
    \min_{\theta,\phi,\mW_q,\mW_k}
    \Ls_{\text{cls}} + \lambda_{\text{aug}} \Ls_{\text{aug}}
\end{equation}
where $\lambda_{\text{aug}}$ denotes the trade-off parameter.

\subsection{Adversarial Feature Memory Bank}
As mentioned earlier, the feature vectors $\{\vz_i\}$ in the memory bank $\tZ$ act as basis vectors that linearly span the subspace, $\text{span}(\tZ)$. To be effective, these vectors must be both representative and diverse so that they can create discriminative and generalizable 
augmenting feature vectors $\vz^a$. We now present our proposed method for adversarial feature memory initialization and updating.

\subsubsection{Memory Bank Initialization}
The feature memory bank, $\tZ$, is constructed from historical training feature vectors. During the early stages of the training process, the extracted features tend to be unstable. To address this, we perform a warm-up training using only the loss function $\Ls_{\text{cls}}$ for a few iterations.
Once the training stabilizes, we extract feature vectors from the training set to populate the memory bank. However, randomly selecting features for memory bank initialization does not guarantee the diversity needed for an effective memory bank. One possible approach is to uniformly sample an equal number of feature vectors from each category. However, because the features may not yet be sufficiently discriminative at this early stage, feature vectors from different categories might still be quite similar.

To enhance diversity, we propose applying K-Means clustering \citep{arthur2006k} to all extracted feature vectors and randomly selecting vectors from each cluster based on cluster size. Assuming a total of $K$ clusters, we first perform K-Means clustering to obtain $K$ clusters:
\begin{equation}\label{eq:kmeans}
	\{\tZ_1, \cdots, \tZ_K\} = \text{K-Means}(\{\vz|\vz=f(\vx), \vx\in D\}) 
\end{equation}
When sampling feature vectors from each cluster, a straightforward approach is to sample an equal number of vectors from each cluster, i.e., $N_m/K$. However, the size of each cluster, $|\tZ_k|$, may not always be greater than $N_m/K$. To address this, we propose sampling feature vectors from each cluster in proportion to the cluster size:
\begin{equation}\label{eq:randompick}
    \tZ = \cup_{i=1}^K \text{RandomPick}(\tZ_k, N_m\times |\tZ_k|/|D|)
\end{equation}
Here, $\text{RandomPick}(\tZ_k, N_m\times |\tZ_k|/|D|)$ represents the random sampling of $N_m\times |\tZ_k|/|D|$ vectors from the cluster $\tZ_k$, with $N_m$ representing the size of the memory bank. By combining the sampled feature vectors from each category, we create a diverse initial memory bank.

\subsubsection{Adversarial Memory Bank Update}
After the warm-up stage of the training process, the feature memory bank is updated to incorporate the most recent training features, ensuring that the query feature $\vz$ is comparable with the memory features $\vz_i$. However, because the training features are confined to the training domain's distribution, they may not be sufficiently representative or generalizable to unseen testing domains.
To address this, we propose an adversarial feature generation method that creates features beyond the training domain distribution for updating the memory bank. Specifically, for an instance $\vx$ sampled from the training set $D$, the feature encoder $f_\theta$ extracts the feature vector $\vz$, which serves as the starting point ($\vz^1 = \vz$). We then apply Langevin dynamics-based diffusion \citep{braun2024deep} to iteratively update the feature vector using noisy gradients for $T$ iterations. At each time step $t$, the feature vector is updated as follows:
\begin{equation}\label{eq:adversarial-feature-concat}
    \vz^{t+1} = \vz^{t} + \eta(t) \nabla_{\vz^t} \ell_{ce}(h_\phi(\vz^t; \vz^t), \vy) + \sqrt{2\eta(t)} \rvepsilon^t
\end{equation}
where $\rvepsilon^t$ is an isotropic Gaussian random vector with zero mean and unit variance, i.e., $\rvepsilon^t \sim \gN(\mathbf{0}, \mI_{d_\vz})$ for $t \in [1, T]$. The coefficient $\eta(t) = \eta_0 / (t + 1)$ represents the decreasing step size.
Considering feature duplication, the gradient of $\ell_{ce}$ with respect to the feature vector $\vz^t$ can be analyzed from the logits-level ensemble perspective described earlier. With the expanded weight matrix $\mW = [\mW_1, \mW_2]$ of the linear classification head $h_\phi$, the logits vector $h_\phi([\vz^t; \vz^t])$ is computed as $(\mW_1 + \mW_2) \vz^t$. Thus, the gradient can be easily computed and is also solvable using automatic differentiation tools, such as PyTorch.

As gradient ascent increases the loss value, the adversarial feature updating procedure generates feature vectors outside the training feature distribution, thereby enhancing the representativeness of the memory bank. 
During each memory bank update step, we replace a portion of the feature vectors in the memory bank with newly generated adversarial features, with a hyperparameter $\gamma$ (e.g., 70\%) controlling the update ratio. To avoid losing representative vectors for better generalization, 
we remove less informative vectors from the memory bank based on entropy.
Specifically, for each memory feature vector $\vz_i$, we duplicate it and calculate the prediction probability vector $\vp_i$ from the expanded classification head $h_\phi$, i.e., $\vp_i = \text{softmax}(h_\phi([\vz_i; \vz_i]))$. The entropy of each predicted probability vector $\vp_i$ is computed as $-\vp_i^\top \log \vp_i$. Feature vectors with lower entropy values are removed from the memory bank, and the newly generated adversarial feature vectors are used to fill the vacant spaces in the bank.
The overall training algorithm for the proposed method, SDGAM, is presented in Algorithm~\ref{alg:SDGAM}.

\begin{algorithm}[]
\caption{Single Domain Generalization with Adversarial Memory (SDGAM)}\label{alg}
	\label{alg:SDGAM}
\SetKwInOut{Input}{Input}\SetKwInOut{Output}{Output}
\Input{Classification model $f_\theta\circ h_\phi$, projection matrices $\mW_q$ and $\mW_k$, training data $D$, number of epochs $E$.}
Warm-up training $f_\theta$ and $h_\phi$ using $\Ls_{\text{cls}}$;\\
Initialize memory bank $\tZ$;\\
\For{epoch $\leftarrow 1$ \KwTo $E$}{
    Generate adversarial features, update memory;\\
    \For{batch $\{x\}\subset D$}{
        Extract features $\vz=f_\theta(\vx)$;\\
        Generate augmenting features $\vz^a=g(\vz)$;\\
        Calculate loss $\Ls_{\text{cls}}+\lambda_{\text{aug}} \Ls_{\text{aug}}$;\\
        Update model parameters $\theta, \phi, \mW_q$ and $\mW_k$;\\
    }
}
\Output{$f_\theta\circ h_\phi$, $\mW_q$, $\mW_k$ and $\tZ$.}
\end{algorithm}

\subsection{Inference with Memory}

For inference on testing data, we use the classification model $f \circ h$ and the feature augmentation network $g(\cdot; \mW_q, \mW_k, \tZ)$. Given an instance $\vx$ sampled from the testing set $T$, we predict the category as follows:
\begin{equation}
    \hat{c} = \argmax_c [h_\phi([f_\theta(\vx); g(f_\theta(\vx); \mW_q, \mW_k, \tZ)])]_c
\end{equation}
Compared to Empirical Risk Minimization (ERM) training, the model for inference with memory includes additional parameters: the expanded classification head $h$, the projection matrices $\mW_q$ and $\mW_k$, and the memory bank $\tZ$. The extra computational cost introduced by the former two components is negligible. The computational cost associated with the memory bank $\tZ$ depends on the bank size $N_m$. Fortunately, $N_m$ is relatively small, as will be demonstrated in the experimental results.

\section{Experiments}
\subsection{Experimental Setup}

\label{section:experimental-setup}
\paragraph{Datasets}
Our experimental evaluation utilizes three pivotal benchmark datasets. For the task of digit classification, we employ the Digits benchmark, comprising five distinct datasets: MNIST \citep{lecun1998gradient}, MNIST-M \citep{ganin2015unsupervised}, SVHN \citep{netzer2011reading}, SYN \citep{ganin2015unsupervised}, and USPS \citep{denker1989neural}. These datasets collectively encompass the same categories, specifically the digits from 0 to 9. In our experimental framework, MNIST is utilized as the training domain, whereas the other datasets serve as testing domains for our evaluations.
We also incorporate the PACS dataset \citep{li2017deeper}, which contains four distinct artistic domains: Art, Cartoon, Photo, and Sketch. Each domain includes the same seven categories of objects. For our purposes, the Photo is selected as the training domain, with the remaining domains acting as testing domains.
Finally, the DomainNet dataset \citep{peng2019moment}, which represents the most rigorous dataset in our study, spans six diverse domains: Real, Infograph, Clipart, Painting, Quickdraw, and Sketch, featuring a comprehensive assortment of 345 object classes. In this case, the Real is designated as the training domain, and the extensive class diversity and domain variability of the remaining domains pose a substantial challenge in our evaluation.
\paragraph{Experimental Setup} 
In our experiments, consistent with previous research \citep{wang2021learning}, we employed LeNet as the primary architecture for the Digits dataset, training it on the initial set of 10,000 MNIST images. All images were standardized to a resolution of $32 \times 32$ and transformed into RGB format. Our experimental configuration consisted of 50 epochs, a batch size of 32, and an initial learning rate of $1e^{-4}$, which was decreased by a factor of 0.1 after 25 epochs.
For the PACS dataset, we fine-tuned a pre-trained ResNet-18, originally trained on ImageNet, on the training domain with images resized to $224 \times 224$. This setup also included 50 epochs, a batch size of 32, and an initial learning rate of 0.001, which was modulated according to a cosine annealing scheduler.
The same ResNet-18 architecture was applied to the DomainNet dataset, but with the experiments extending over 200 epochs and a larger batch size of 128. The learning rate adjustment followed a similar cosine annealing pattern. To ensure statistical reliability, all experiments were replicated five times using different random seeds, and the results were reported as the average accuracy and standard deviation. Specifically for 
the hyperparameters,
we set $\lambda_{\text{aug}}$, $\gamma$, $\beta$ and $r_m$ ($N_m/|D|$) to 1, 0.7, 0.5 and 0.1 respectively. 
\begin{table*}[t]
\centering
\begin{tabular}{l|cccc|c}
\hline
Method & SVHN & MNIST-M & SYN & USPS & Avg. \\ \hline
 ERM \citep{koltchinskii2011oracle} & 27.8 & 52.7 & 39.7 & 76.9 & 49.3 \\
 CCSA \citep{motiian2017unified}& 25.9 & 49.3 & 37.3 & 83.7 & 49.1\\
JiGen \citep{carlucci2019domain} & 33.8 & 57.8 & 43.8 & 77.2 & 53.1\\
ADA \citep{volpi2018generalizing}& 35.5 & 60.4 & 45.3 & 77.3 & 54.6\\

 ME-ADA \citep{Long2020Maximum} & 42.6 & 63.3 & 50.4 & 81.0 & 59.3\\ 
M-ADA \citep{qiao2020learning} & 42.6 & 67.9 & 49.0& 78.5 & 59.5\\
AutoAug \citep{cubuk2018autoaugment} & 45.2 & 60.5 & 64.5 & 80.6 & 62.7 \\
RandAug \citep{cubuk2020randaugment} & 54.8 & 74.0 & 59.6 & 77.3 & 66.4 \\
   RSDA \citep{volpi2019addressing} & {47.7}& 81.5 & 62.0 & 83.1 & 68.5\\
 L2D \citep{wang2021learning}& 62.9 &  \textbf{87.3} & 63.7 & 84.0 & 74.5\\ 
  PDEN \citep{li2021progressive} & 62.2 & 82.2 & 69.4 & 85.3 & 74.8\\
  SimDE \citep{chen2023meta}& 66.0& 84.9& 70.0 & 86.5 &76.8 \\
    MCL \citep{chen2023meta} & 69.9 & 78.3 & 78.4 & 88.5 & 78.8 \\
 \hline 
	 SDGAM (Ours)&$\mathbf{70.2}_{(0.8)}$&${79.8}_{(0.6)}$&$\mathbf{82.8}_{(0.3)}$&$\mathbf{89.1}_{(0.6)}$&$\mathbf{80.5}$
 \\ \hline
\end{tabular}
\caption{Classification accuracy and standard deviation(\%) results on the four testing domains SVHN, MNIST-M, SYN, and USPS, with MNIST as the training domain. Results are shown as $\text{mean}_{(\text{std})}$ with best results in bold font.}
\label{tab:digits}
\end{table*}
\begin{table*}[t]
\centering
\setlength{\tabcolsep}{1mm}
\begin{tabular}{c|cccccccccc|c}
\hline
Testing & MixUp & CutOut & ADA & ME-ADA & AugMix & RandAug & ACVC& L2D  &SimDE& MCL &SDGAM (Ours) \\ \hline
Art & 52.8 & 59.8 & 58.0 & 60.7 & 63.9 & 67.8 & 67.8 & 67.6 &-&-&$\mathbf{69.7}_{(0.8)}$ \\
Cartoon & 17.0 & 21.6 & 25.3 & 28.5 & 27.7 & 28.9 & 30.3 & 42.6 &-&-&$\mathbf{53.9}_{(0.9)}$ \\
Sketch & 23.2 & 28.8 & 30.1 & 29.6 & 30.9 & 37.0 & 46.4 & 47.1 &-&-&$\mathbf{56.3}_{(0.8)}$ \\
\hline
Avg. & 31.0 & 36.7 & 37.8 & 39.6 & 40.8 & 44.6 & 48.2 & 52.5 &$59.3$ &59.6 &$\mathbf{59.9}$ \\ 
\hline
\end{tabular}
\caption{Classification accuracy and standard deviation(\%) results on the PACS dataset. Photo is used as the training domain.}
\label{tab:pacs-experiment}
\end{table*}
\begin{table*}[t]
\centering
\setlength{\tabcolsep}{1mm}
\begin{tabular}{c|ccccccccc|c}
\hline
Testing    & MixUp & CutOut & CutMix & ADA  & ME-ADA & RandAug & AugMix & ACVC &SimDE         & SDGAM (Ours)                                        \\ \hline
Painting  & 38.6  & 38.3   & 38.3   & 38.2 & 39.0   & 41.3    & 40.8   & 43.6       &39.9   & $\mathbf{43.8}_{(0.8)}$ \\
Infograph & 13.9  & 13.7   & 13.5   & 13.8 & 14.0   & 13.6    & 13.9   & 12.9       &12.9   & $\mathbf{14.8}_{(0.9)}$  \\
Clipart   & 38.0  & 38.4   & 38.7   & 40.2 & 41.0   & 41.1    & 41.7   & ${42.8}$ & 41.7&$\mathbf{43.5}_{(0.7)} $ \\
Sketch    & 26.0  & 26.2   & 26.9   & 24.8 & 25.3   & 30.4    & 29.8   & 30.9       &33.4   & $\mathbf{34.5}_{(0.8)}$ \\
Quickdraw & 3.7   & 3.7    & 3.6    & 4.3  & 4.3    & 5.3     & 6.3    & 6.6  &6.8& $\mathbf{7.2}_{(1.0)}$    \\
\hline
Avg.      & 24.0  & 24.1   & 24.2   & 24.3 & 24.7   & 26.3    & 26.5   & 26.9    &26.9      & $\mathbf{28.7}$ \\ 
\hline
\end{tabular}
\caption[DomainNet result]{Classification accuracy and standard deviation(\%) results on the DomainNet dataset. 
Real serves as the training domain.}
\label{tab:domainnet}
\end{table*}

\subsection{Comparison Results}
We compare our method with several existing methods including MixUp \citep{zhang2018mixup}, CutOut \citep{devries2017improved}, 
CutMix \citep{yun2019cutmix}, 
AugMix \citep{hendrycks2019augmix}, ACVC \citep{Cugu_2022_CVPR}, ERM \citep{koltchinskii2011oracle}, CCSA \citep{motiian2017unified}, JiGen \citep{carlucci2019domain}, ADA \citep{volpi2018generalizing}, ME-ADA \citep{Long2020Maximum}, M-ADA \citep{qiao2020learning}, AutoAug \citep{cubuk2018autoaugment}, RandAug \citep{cubuk2020randaugment}, RSDA \citep{volpi2019addressing}, L2D \citep{wang2021learning}, PDEN \citep{li2021progressive}, SimDE \citep{xu2023simde},  and MCL \citep{chen2023meta}.

Table \ref{tab:digits} presents the comparative results on the digits dataset using LeNet as the backbone network. In our evaluation, the SDGAM method demonstrated consistent and noteworthy improvements in terms of percentage gains over the MCL method across the testing domains. Specifically a significant enhancement was observed in the SYN domain, where SDGAM outperformed MCL, the second-best approach, by 4.40\%, showcasing its robust capacity to manage extensive domain shifts. On average, the improvement over MCL stood at 1.70\%, highlighting SDGAM's refined ability to handle domain-specific characteristics efficiently. These improvements underscore SDGAM's enhanced generalization capabilities across varied and challenging datasets.

Table \ref{tab:pacs-experiment} reports the comparison results on the PACS dataset with ResNet-18 as the backbone network. In the PACS dataset comparison, our SDGAM method showcased consistent
improvements over the previously leading approach, MCL. 
On average, SDGAM achieved a top score of 59.9\%,  outperforming MCL's 59.6\%. This indicates a consistent edge across diverse artistic domains, affirming SDGAM's superior adaptability and generalization capability in handling domain shifts within the PACS dataset.

Table \ref{tab:domainnet} reports the comparison results on the DomainNet dataset with ResNet-18 as the backbone network. In the DomainNet dataset comparison, our SDGAM method consistently surpassed other approaches, demonstrating its effectiveness across a range of diverse domains. Notably, SDGAM presented clear percentage improvements over the second-best results, highlighting its refined adaptation capabilities.
Notablly for the Infograph domain, SDGAM is marking a clear improvement of 1.9\% over the state-of-the are SimDE, underscoring its effectiveness in dealing with complex, information-dense images.
On average, SDGAM achieved an overall top score of 28.7\%,  outperforming the closest contenders ACVC and SimDE, both at 26.9\% by a significant margin of 1.8\%. This performance confirms SDGAM's superior adaptability and its robust generalization capabilities across a spectrum of visually diverse and challenging domains within the DomainNet dataset.

Overall, the proposed SDGAM framework consistently outperforms existing single domain generalization methods across various experimental settings, demonstrating its exceptional effectiveness, particularly in handling significant domain shifts between training and testing domains.

\subsection{Ablation Study}

\begin{table}[t]
\centering
\setlength{\tabcolsep}{1.8mm}
\begin{tabular}{l|c|c|c|c|c}
\hline
Method & SVHN & MM & SYN & USPS & Avg. \\ \hline
        SDGAM (Ours)&$\mathbf{70.2}$&$\mathbf{79.8}$&$\mathbf{82.8}$&$\mathbf{89.1}$&$\mathbf{80.5}$\\\hline
        $- \text{w/o } \mathcal{L}_\text{aug}$ &${30.1}$ & ${53.1}$  & ${42.3}$  & ${81.3}$  & ${51.7}$ \\
                   $- \text{w/o  Advers.}$ &${68.1}$ & ${76.2}$  & ${75.7}$  & ${85.9}$  & $77.7$ \\
                          $- \text{w/o  Concat}$ &${65.2}$ & ${50.9}$  & ${45.4}$  & ${82.0}$  & 60.8 \\
                            $- \text{w/o  t-Mem.}$ &${66.3}$ & ${75.2}$  & ${78.9}$  & ${85.0}$  & 76.3 \\
                 
                   \hline

\end{tabular}
\caption{Ablation Study Classification accuracy and standard deviation(\%) results on the four testing domains SVHN, MM (MNIST-M), SYN, and USPS, with MNIST as the training domain. Best results are in bold font.}
\label{tab:ablation}
\end{table}
In order to investigate the contribution of each component of the proposed framework, we conducted an ablation study to compare the proposed SDGAM with its four
variants: 
(1) ``$\; - \text{w/o } \mathcal{L}_{\text{aug}}$ ", which drops the augmentation loss
and is equivalent to ERM training;
(2) ``$\; - \text{w/o } $ Advers.",  
which 
drops adversarial process in Eq. \ref{eq:adversarial-feature-concat};
(3) ``$\; - \text{w/o } $ Concat",
which
excludes the concatenation step, thereby examining the effect of 
directly using augmenting features $\vz^a$ for classification;
(4) ``$\; - \text{w/o }$ t-Mem.",
which drops accessing the memory module during test phase evaluation, relying solely on the primary neural network architecture for predictions.

The ablation study outlined in Table \ref{tab:ablation} provides a comprehensive analysis of the individual contributions of various components in the 
proposed SDGAM approach. The results underscore the significant impact each component has on classification accuracy across four testing domains (SVHN, MNIST-M, SYN, and USPS) with MNIST as the training domain.
The removal of the feature augmentation loss, $\mathcal{L}_\text{aug}$, results in the most drastic performance drop. This indicates that $\mathcal{L}_\text{aug}$ is critical for enhancing the model's generalization capability
as it implicitly aligns the training and testing domain distributions in the projected space.
Excluding the adversarial process in Eq.\ref{eq:adversarial-feature-concat} also leads to a noticeable decrease in performance across all domains. This suggests that 
adversarial feature generation plays a substantial role in creating representative and generalizable memory features.
The removal of the concatenation operation shows a significant impact, particularly in the MNIST-M and SYN domains, where accuracies drop to 50.9\% and 45.4\%, respectively. This component might be pivotal in effectively merging style and semantic features, which is essential for maintaining performance consistency across varied domains.
Lastly, conducting the test evaluation without the memory component ($\text{w/o t-Mem.}$) results in a lesser, yet significant decline in average accuracy, dropping to 76.3\%. This emphasizes the memory's critical role in 
aligning the feature distributions in the projected space
during the inference phase, thereby enhancing both adaptability and accuracy in previously unseen domains.
Overall, the ablation study clearly demonstrates the necessity of each component within the SDGAM framework, with each element playing a distinct and crucial role in strengthening the model's ability to effectively generalize across significantly different domains.
\subsection{Hyperparameter Selection}
\begin{figure*}[t]
\centering
\begin{subfigure}{0.24 \textwidth}
\centering

\includegraphics[width = \textwidth, height=1.15in]{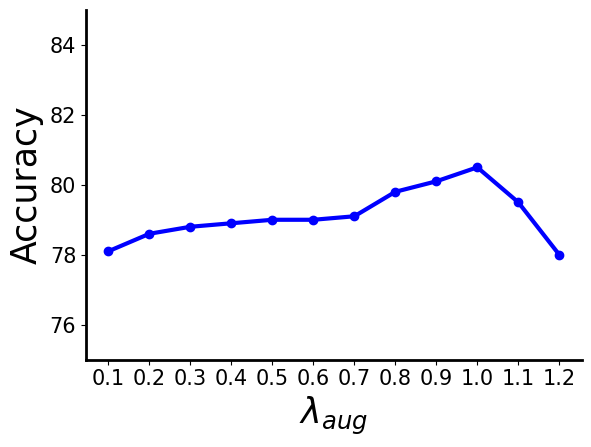}
\caption{$\lambda_{\text{aug}}$}
\end{subfigure}
\begin{subfigure}{0.24\textwidth}
\centering
\includegraphics[width = \textwidth, height=1.15in]{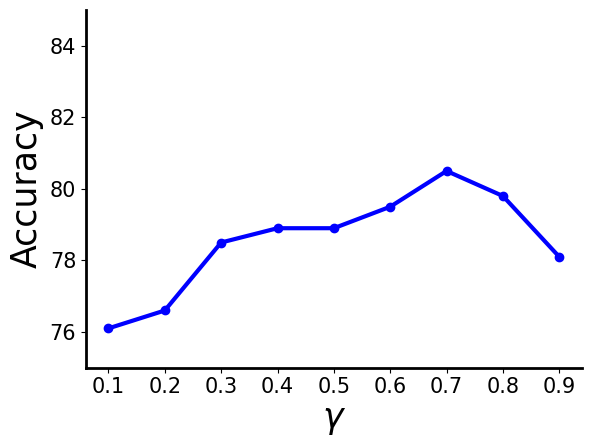}

\caption{$\gamma$}
\end{subfigure}
\begin{subfigure}{0.24\textwidth}
\centering
\includegraphics[width = \textwidth, height=1.15in]{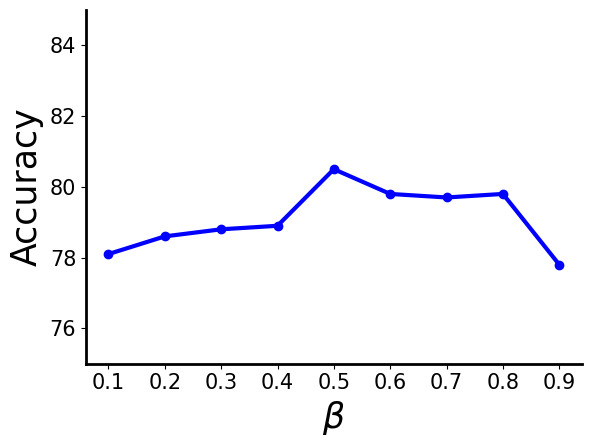}

\caption{$\beta$}
\end{subfigure}
\begin{subfigure}{0.24\textwidth}

\centering
\includegraphics[width = \textwidth, height=1.15in]{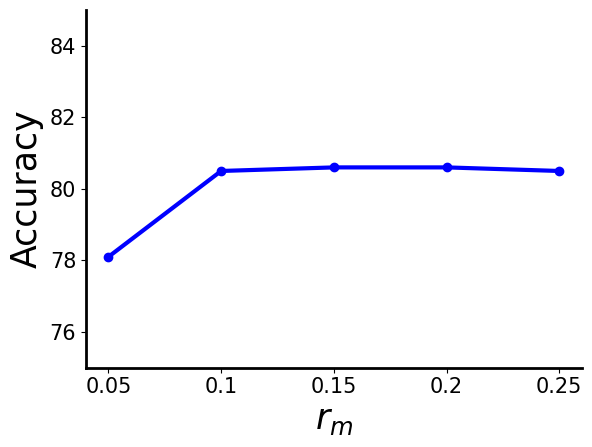}

\caption{$r_m$}
\end{subfigure}
	\caption{Hyperparameter selection analysis for four hyper-parameters $\lambda_{\text{aug}}$, $\gamma$,  $\beta$, and $r_m$ ($N_m/|D|$) on digits dataset.}

\label{fig:hyper_sen}
\end{figure*}

We perform a comprehensive hyperparameter selection analysis of the SDGAM framework across four critical hyperparameters: $\lambda_{\text{aug}}$, the trade-off parameter for the feature augmentation
loss term; $\gamma$, the memory update rate; $\beta$ the trade-off parameter for label vector combination in Eq~\ref{eq:label-mixup}; and $r_m$, the memory size ratio relative to the entire training dataset (i.e $N_m/|D|$).
We executed our experiments on the Digits dataset, independently varying each of the four hyperparameters across a range of values. The results  are presented in Figure \ref{fig:hyper_sen}. 

The hyperparameter  $\lambda_{\text{aug}}$ was evaluated by varying its value from 0.1 to 1.2. The results indicate a progressive improvement in performance with an increase in $\lambda_{\text{aug}}$ up to $\lambda_{\text{aug}}=1$.  This trend suggests that a higher weight on the memory loss term enhances the model's ability to retain and utilize past information effectively, thus improving overall performance. However, overly large values for $\lambda_{\text{aug}}$ can degrade the model by overshadowing the labeled loss effect.
The analysis for $\gamma$ revealed a noticeable influence on performance, with values ranging from 0.1 to 0.9. The results indicate that the accuracy increased from 76.1\% at $\gamma = 0.1$ to a maximum of 80.5\% at $\gamma = 0.7$. Beyond this point, the performance decreased slightly, reaching 78.1\% at $\gamma = 0.9$. This pattern suggests that an appropriate rate of memory update enhances the model's adaptability and learning efficiency. However, overly frequent updates may introduce noise and destabilize the learning process, thus decreasing performance.
The impact of $\beta$ on the framework's performance was assessed by altering its value from 0.1 to 0.9. The performance initially increased, peaking at $\gamma = 0.5$. However, further increases in $\beta$ led to a decline in performance. This behavior can be attributed to the trade-off between the combination of primary and memory labels. At moderate levels, the balance is optimal, but excessive focus on label combination can distract the model from the primary task, leading to reduced accuracy.
The memory size ratio, $r_m$, was varied between 5\% and 25\% of the training data. The analysis showed a significant improvement in accuracy from 78.1\% at $r_m = 5\%$ to 80.6\% at $r_m = 10\%$. This trend indicates that a too-small memory size limits the model's generalization ability, 
while an excessively large memory size can lead to redundancy and increased computational burden.

\section{Conclusion}
This paper addresses the single domain generalization problem by introducing SDGAM, a method that leverages an adversarial feature memory bank to augment training features. The feature vectors stored in the memory bank serve as key and value vectors, which are used to compute attention vectors for the query feature vectors. The proposed feature augmentation network utilizes these attention vectors and memory features to map the extracted training features into an invariant subspace spanned by the memory features. This approach implicitly aligns the training and testing domains in the projected space without requiring access to unseen testing data.
To maintain the diversity of the memory bank, we introduce an adversarial feature generation method that updates the memory features using noisy gradients, generating features that extend beyond the training domain distribution. 
Experimental results show that SDGAM achieves state-of-the-art performance on standard single domain generalization benchmarks.

\bibliography{reference}
\end{document}